\documentclass[letterpaper,10pt,conference]{ieeeconf}

\IEEEoverridecommandlockouts
\usepackage{xparse} 
\usepackage{cite}
\usepackage{amsmath,amssymb,amsfonts}
\usepackage{algorithmic}
\usepackage{graphicx}
\graphicspath{{imgs/}}
\usepackage{textcomp}
\usepackage{xcolor}
\usepackage{hyperref}
\usepackage{array}
\usepackage{multicol}
\usepackage{multirow}
\usepackage{booktabs}
\usepackage{tablefootnote}
\usepackage{color,colortbl}
\usepackage[normalem]{ulem}
\IEEEtriggeratref{23}

\title{\Large\bf Learning to Detect Slip with Barometric Tactile Sensors\\ and a Temporal Convolutional Neural Network}

\author{Abhinav Grover$^\ast$, Philippe Nadeau$^\ast$, Christopher Grebe, and Jonathan Kelly$^\dagger$
\thanks{All authors are with the Space \& Terrestrial Autonomous Robotics Systems (STARS) Laboratory at the University of Toronto Institute for Aerospace Studies, Toronto, Ontario, Canada. {\tt\footnotesize <firstname>.<lastname>@robotics.utias.utoronto.ca}}
\thanks{$^\ast$Abhinav Grover and Philippe Nadeau were supported in part by the Vector Institute Scholarship in Artificial Intelligence.}
\thanks{$^\dagger$Jonathan Kelly is a Vector Institute Faculty Affiliate. This research was supported in part by the Canada Research Chairs program.}}

\begin{document}
\maketitle

%%%%%%%%%%%%%%%%%%%%%%%%%%%%%%%%%%%%%%%%%%%%%%%%%%%%%%%%%%%%%%%%%%%%%%%%%%%%%%%%
%% ABSTRACT
%%%%%%%%%%%%%%%%%%%%%%%%%%%%%%%%%%%%%%%%%%%%%%%%%%%%%%%%%%%%%%%%%%%%%%%%%%%%%%%%
\begin{abstract}
The ability to perceive object slip via tactile feedback enables humans to accomplish complex manipulation tasks including maintaining a stable grasp.
Despite the utility of tactile information for many applications, tactile sensors have yet to be widely deployed in industrial robotics settings; part of the challenge lies in identifying slip and other events from the tactile data stream.
In this paper, we present a learning-based method to detect slip using barometric tactile sensors.
These sensors have many desirable properties including high durability and reliability, and are built from inexpensive, off-the-shelf components.
We train a temporal convolution neural network to detect slip, achieving high detection accuracies while displaying robustness to the speed and direction of the slip motion.
Further, we test our detector on two manipulation tasks involving a variety of common objects and demonstrate successful generalization to real-world scenarios not seen during training.
We argue that barometric tactile sensing technology, combined with data-driven learning, is suitable for many manipulation tasks such as slip compensation.
\end{abstract}

%%%%%%%%%%%%%%%%%%%%%%%%%%%%%%%%%%%%%%%%%%%%%%%%%%%%%%%%%%%%%%%%%%%%%%%%%%%%%%%%
%% INTRODUCTION
%%%%%%%%%%%%%%%%%%%%%%%%%%%%%%%%%%%%%%%%%%%%%%%%%%%%%%%%%%%%%%%%%%%%%%%%%%%%%%%%
\section{Introduction}

Tactile signals provide vital information about slip, relative motion at the contact interface between the hand and an object, faster than any exteroceptive perception method \cite{westling1984factors,johansson1984roles}.
Slip can be disastrous (e.g., when transporting a fragile object) or advantageous (e.g., when pushing an object without lifting it) depending on the context \cite{wang2020swingbot}.
In robotics, the well-studied `handover' task, which involves passing an object from a robot hand to a human hand, requires control of the gripping force with accuracy and speed to avoid significant slip \cite{ortenzi2020object}.
The requisite feedback can only be provided through tactile sensing \cite{chan2012grip} and, consequently, the detection and control of slip is fundamental to the completion of handovers and many other tasks.
Despite significant recent progress, artificial tactile sensors have yet to achieve the fidelity or accuracy of human tactile perception.

A wide range of new tactile sensors have become available over the last decade. 
These sensors measure various physical properties including capacitance \cite{maslyczyk2017highly}, impedance \cite{johansson2011biomimetic}, or optical changes \cite{lambeta2020digit,donlon2018gelslim,johnson2011microgeometry,chorley2009development,johansson2011biomimetic}.
Each tactile sensor type has inherent characteristics such as fragility, bulk, resolution, nonlinearity, hysteresis, and cost. 
Consideration of these characteristics and how they relate to varying task constraints leads to the best choice of tactile sensor for a specific job. 

\begin{figure}[t]
\vspace{1mm}
\centering
\includegraphics[width=0.95\columnwidth]{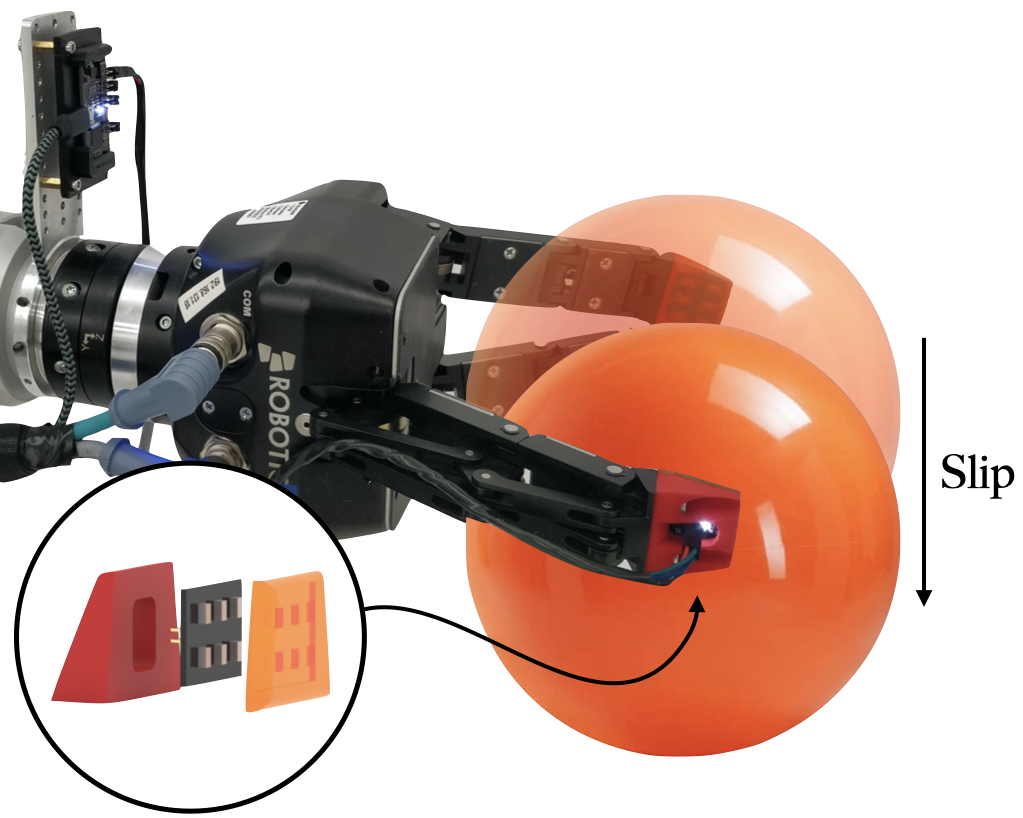}
    \caption{We perform in-hand slip detection using barometric tactile sensors. Our approach involves training a temporal convolutional neural network (TCN) to recognize slip events directly from tactile data stream. The tactile sensors are mounted on the fingertips of a Robotiq 3-finger adaptive gripper. Inset: exploded view of each RightHand Labs TakkTile sensor (plastic support frame, barometer circuit board, and rubber matrix, from left to right).}
    \label{fig:actution_hardware}
    \vspace{-5mm}
\end{figure}

In this paper, we investigate the potential of low-cost barometric tactile sensors, in combination with a state-of-the-art neural network, to perform slip detection.
The sensors (the TakkTile model from RightHand Labs) are built from an array of commercial MEMS barometers fixed to a PCB backplane, with a thin rubber matrix forming the contact surface.
%
% In bulk, the fabrication cost to make an array similar to the one used in this work is about \$15 USD.
%
These sensors have a low profile, are mechanically robust, exhibit a consistent linear pressure response, and are easy to integrate with existing end-effectors.
The complex spatiotemporal signature of pressure changes during slip is difficult to model analytically. Instead, we take a data-driven approach by training a temporal convolutional neural network (TCN) to classify the time-series data produced by the tactile sensors as either static or slipping.
Our contributions are the following:
\begin{itemize}
	\item an algorithm for slip detection using low-cost barometric sensors that achieves an accuracy of over 91\% on average on our validation dataset;
	\item a comparison of our TCN approach with two other slip-detection methods that rely on vibration data;
	\item a preliminary analysis of the sensitivity and robustness of the TCN detector to factors related to surface properties and slipping motion;
	\item extensive experimental results for in-hand slip detection involving objects with various curvatures, hardnesses, and surface properties.
\end{itemize}

%%%%%%%%%%%%%%%%%%%%%%%%%%%%%%%%%%%%%%%%%%%%%%%%%%%%%%%%%%%%%%%%%%%%%%%%%%%%%%%%
%% RELATED WORK
%%%%%%%%%%%%%%%%%%%%%%%%%%%%%%%%%%%%%%%%%%%%%%%%%%%%%%%%%%%%%%%%%%%%%%%%%%%%%%%%
\section{Related Work}
\label{sec:related_work}

\begin{figure}[b]
    \centering
    \vspace{-3.2mm}
    \includegraphics[width=0.82\columnwidth]{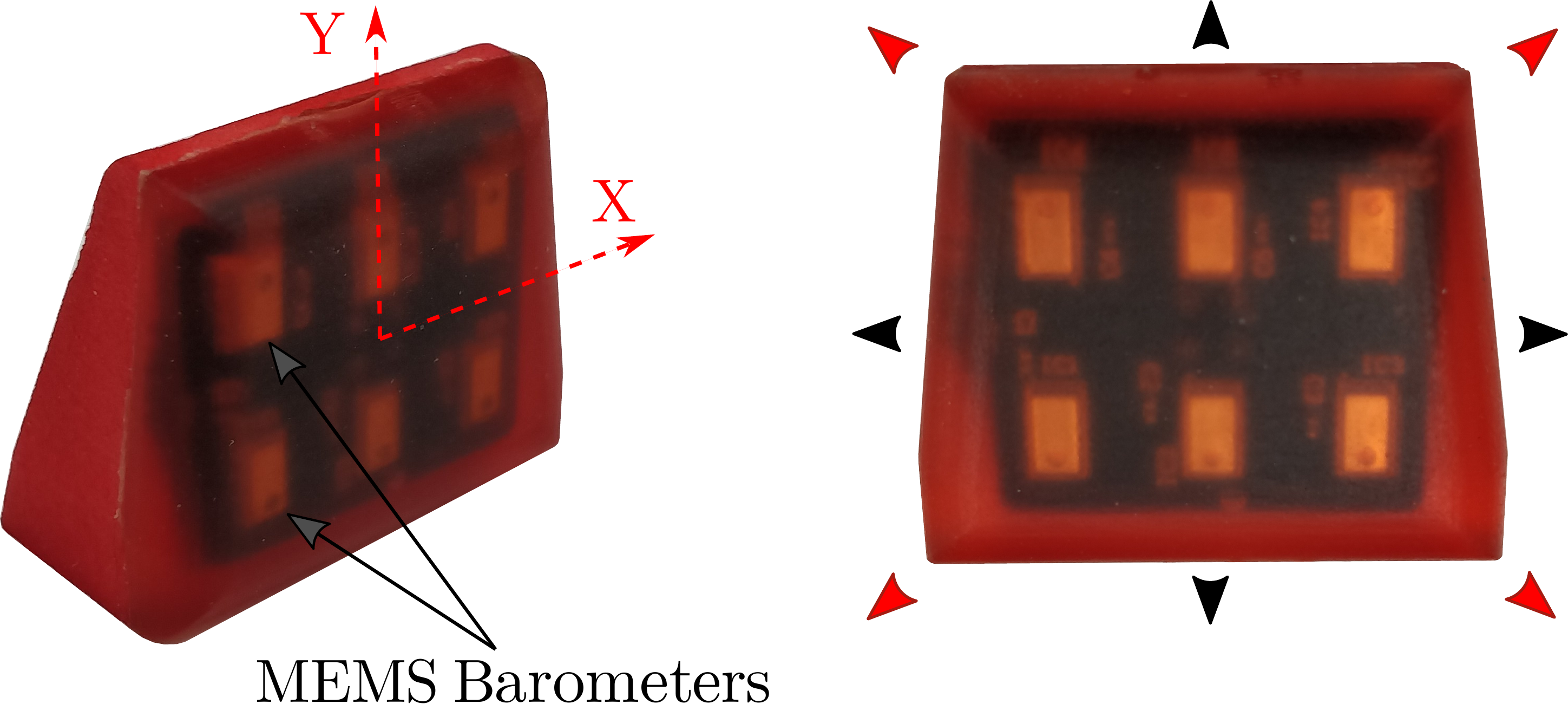}
    \caption{Perspective (left) and front (right) view of the TakkTile fingertip. As part of the front view image, black arrows show the primary sensor axes and red arrows show the oblique axes.}
    \label{fig:TakkTile_sensor}
\end{figure}

\begin{figure}[t]
    \centering
    \includegraphics[width=0.155\textwidth]{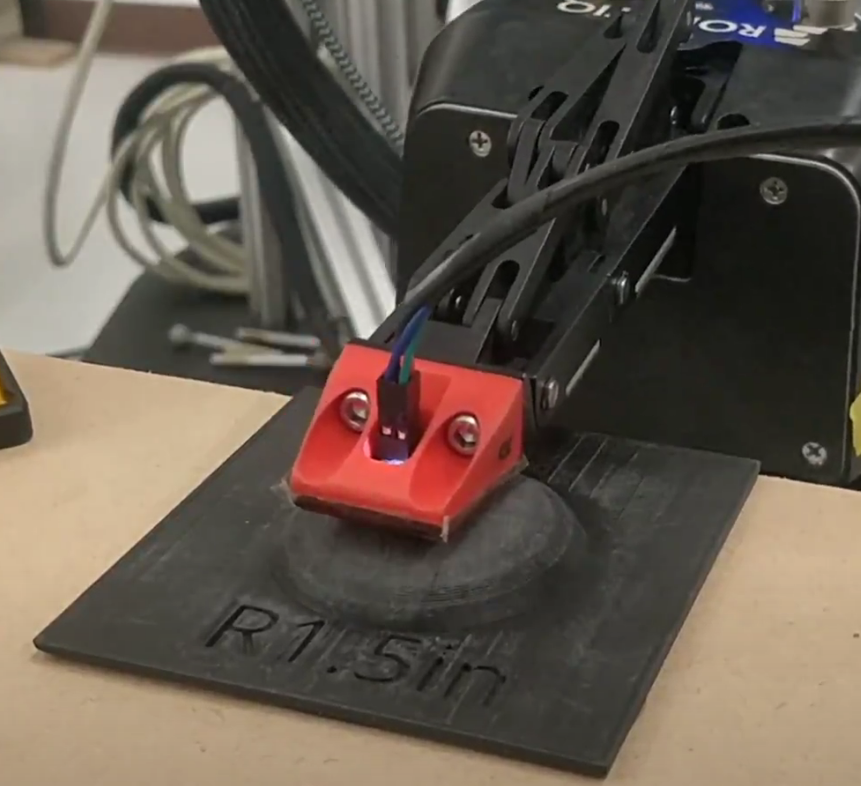}
    \includegraphics[width=0.155\textwidth]{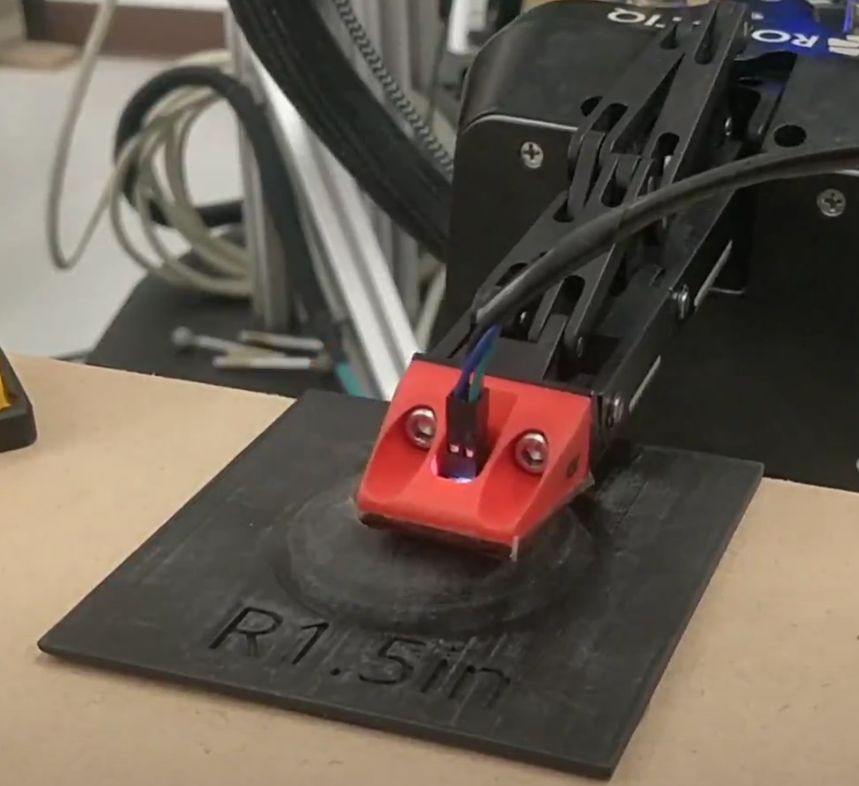}
    \includegraphics[width=0.155\textwidth]{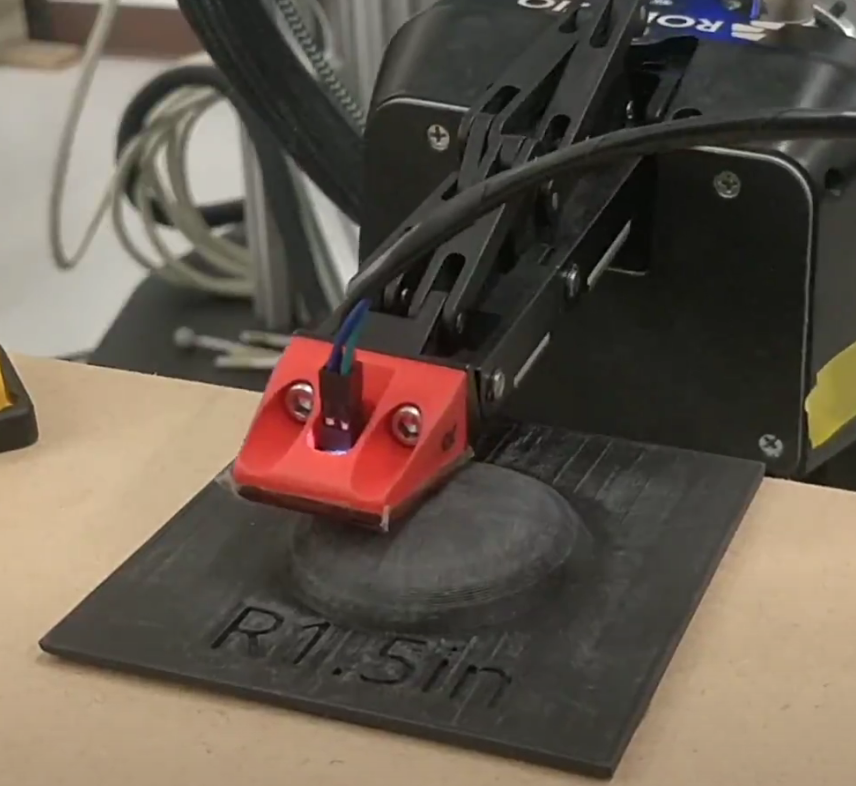}
    \caption{TakkTile sensor mounted on the fingertip of our Robotiq gripper \cite{Tenzer2014} and actuated automatically to slide over the 3D printed spherical surface during training data collection.}
    \label{fig:data_collection}
    \vspace{-4mm}
\end{figure}

\begin{figure}[b]
    \centering
    \includegraphics[width=\columnwidth]{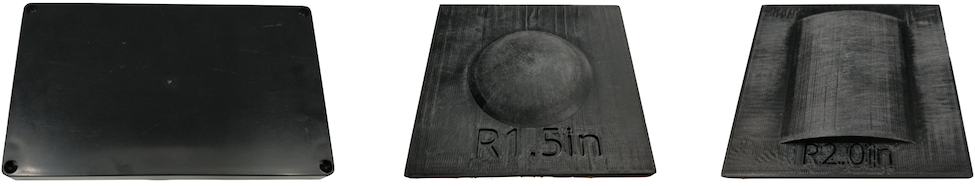}
    \caption{Surfaces used for slip data acquisition, from left to right: a planar box lid, a spherical plastic surface (radius 1.5 in), and a cylindrical plastic surface (radius 2 in). The latter two surfaces were 3D printed and sanded to remove irregular edges.}
    \label{fig:slip_surfaces}
\end{figure}

\begin{figure*}[t]
    \centering
    \includegraphics[width=0.98\textwidth]{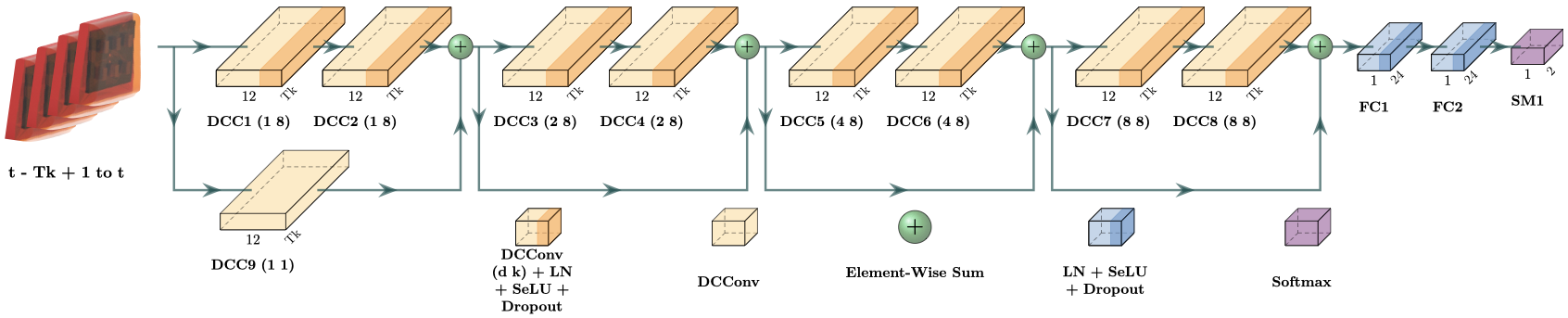}
    \vspace{-2mm}
    \caption{The TCN \cite{Bai2018} architecture used for slip detection. Temporal features are extracted using dilated convolution layers (DCC)\cite{oord2016wavenet}. For each DCC, $d$ represents the dilation and $k$ represents the kernel size. The next two layers are fully connected, followed by a softmax output for classification.}
    \label{fig:TCN_architecture}
    \vspace{-4mm}
\end{figure*}

% Tactile sensing -- reviews and prior pubs.
The study of tactile sensing for robotic systems has an extensive history, stretching back more than four decades (see, e.g., \cite{1984_Fearing_Basic} and \cite{1984_Raibert_All} for pioneering work).
This research effort has been driven, in large part, by an evolving understanding of the essential role played by the sense of touch in human dexterous manipulation \cite{2010_Ravinder_Tactile}.
For a comprehensive review of existing sensors types, see \cite{cutkosky2016force}; a review of slip detection, in turn, is provided by \cite{Romeo2020}.
We focus the remainder of our review below on slip detection using tactile sensors.

Detecting slip is far from trivial. In an effort to make slip events easier to identify, some researchers have introduced  signal-amplifying structures (e.g. rubber nibs or elastic `fingers') or combined several related sensing modalities.
In \cite{melchiorri2000slip}, data from a gripper-mounted tactile sensor are fused with measurements from a force-torque sensor to detect and control slip during object manipulation.
Gorger et al.\ \cite{goeger2009tactile} combine two tactile sensing modalities (piezoelectric and conductive polymer) to generate `tactile features' that are extracted from the fast Fourier transform (FFT) of the resulting signals.
The features are grouped by a clustering algorithm to classify slip events.
In our work, we propose a method that uses a time series of pressure readings only to infer whether an object is slipping, relaxing the need for an expensive force-torque sensor and, to some extent, for high-frequency sampling. 

% Other ways to detect slip be deformation.
Vision-based tactile sensors, which infer tactile information from visually-sensed deformation of the sensor contact surface, are also now popular.
For two well-known vision-based tactile sensors, the GelSight \cite{Yuan_2015} and GelSlim \cite{Dong2018}, slip can be detected by monitoring changes to shear force at the contact surface.
However, as noted in \cite{Zhang2019}, the vision-based GelSight, GelSlim, and also the FingerVision \cite{yamaguchi2017implementing} sensor, can struggle to detect slip in certain situations due to a limited camera frame rate. Cameras typically operate at 30 to 60 frames per second only.
In \cite{james2020slip}, the TacTip sensor is modified to use a camera that operates at up to 120 frames per second so that rapid movements can be better identified. The authors of \cite{james2020slip} train a support vector machine (SVM) classifier for slip detection, which performs with high accuracy for a wide range of slipping speeds.
However, like most other vision-based sensors, this technology is bulky, preventing it from being placed anywhere other than on the fingertips. In contrast, barometric sensors are compact and can be easily distributed on the palm of a hand \cite{koiva2020barometer}.

% Dynamic/frequency/spectrum techniques.
A subset of methods for slip detection rely on an analysis of the vibration pattern induced by object slip, where the frequency of the vibration (due to material resonance) depends on the composition of the surfaces in contact.
A detection approach developed by Holweg et al.\ \cite{holweg1996slip} computes the power spectral density (PSD) of a piezoresistive tactile signal and feeds this data to a neural network to estimate the probability of slip.
In \cite{roberge2016unsupervised}, a dictionary of spectral tactile features is built to encode tactile readings and an SVM is trained to differentiate between object-environment and object-gripper slip.
In Section \ref{sec: comparison} we compare the performance of a frequency-domain detection method to our proposed method.

% Data-driven techniques.
Due to the complex nature of tactile signals, data-driven approaches are increasingly being used for grasp stability assessment and slip detection \cite{hyttinen2015learning,dang2014stable,goeger2009tactile,kwiatkowski2017grasp,Meier2016,Dong2018}.
The output of tactile sensors can be represented by 2D pressure images that are naturally suited as inputs to convolutional neural networks (CNNs). 
For instance, the work of Meier et al.\ \cite{Meier2016} uses a CNN to detect slip events and to differentiate between rotational and translational slip.
Tactile signals also constitute time-series data that can be fed as the input to recurrent neural networks (RNNs) \cite{Veiga2018, Zapata-Impata2019, Zhang2019, nadeau2020tactile,2020_Begalinova_Self-Supervised}.
Experimental results in \cite{Bai2018} suggests that temporal convolutional networks can outperform conventional RNNs for a diverse set of sequence modeling tasks.
This type of architecture has previously been used for other sequence modelling tasks such as audio synthesis \cite{oord2016wavenet} and for inertial data processing \cite{kaufmann2020deep}.
To the best of the authors' knowledge, TCNs have not yet been applied to process the time-series signals from barometric tactile sensors.

%%%%%%%%%%%%%%%%%%%%%%%%%%%%%%%%%%%%%%%%%%%%%%%%%%%%%%%%%%%%%%%%%%%%%%%%%%%%%%%%
%% Tactile Sensor and Data Acquisition
%%%%%%%%%%%%%%%%%%%%%%%%%%%%%%%%%%%%%%%%%%%%%%%%%%%%%%%%%%%%%%%%%%%%%%%%%%%%%%%%
\section{Slip Detection with Barometric Sensors}

Slip is a complex spatiotemporal event that is challenging to detect with any artificial tactile device.  
In this section, we introduce our barometric tactile sensors and describe a slip data acquisition process that ensures data diversity, with the goal of improving generalization.

%%%%%%%%%%%%%%%%%%%%%%%%%%%%%%%%%%%%%%%%%%%%%%%%%%%%%%%%%%%%%%%%%%%%%%%%%%%%%%%%
\subsection{MEMS Barometric Tactile Sensors}
\label{method: sensors}

We use an off-the-shelf TakkTile sensor kit manufactured by RightHand Labs.
The kit includes three fingertip mounts designed to retrofit the Robotiq 3-finger gripper (see Fig.\ \ref{fig:actution_hardware}).
The sensor design is based on the work of Tenzer et al.\ \cite{Tenzer2014}.
An array of commercial MEMS barometers (NXP MPL115A2) are assembled on a PCB and cast in a mold with urethane rubber, creating a flexible medium to transduce contact forces into barometric pressure signals.
The rubber is 10 mm thick; six MEMS barometers form a $2 \times 3$ sensing array spanning the contact region, as shown in Fig.\  \ref{fig:TakkTile_sensor}.
Each TakkTile sensor provides pressure and temperature data at a sampling rate of 100 Hz,  higher than that of most vision-based tactile sensors.
The sensors are compact, robust, and inexpensive, while exhibiting good linearity, no noticeable hysteresis, and a high signal-to-noise ratio \cite{Tenzer2014}. 

%%%%%%%%%%%%%%%%%%%%%%%%%%%%%%%%%%%%%%%%%%%%%%%%%%%%%%%%%%%%%%%%%%%%%%%%%%%%%%%%
\subsection{Slip Data Acquisition}
\label{sec:data-collection}

In order to collect consistent training data, our acquisition setup leverages a UR10 robot arm equipped with the Robotiq 3-finger gripper. Each TakkTile sensor is fitted to a fingertip (see Fig.\ \ref{fig:data_collection}) and the sensor is moved in a series of patterns over several of custom-made surfaces.
We use a force-feedback PID controller to ensure uniform surface contact throughout each motion; the robot's joint encoders are used to determine the velocity of the fingertip. 
The computed fingertip velocity is then applied to label each time step as `stable' or `slipping.'
For practical slip detection, avoiding false negatives (i.e., slip events labelled as stable) is more important than avoiding false positives (i.e., stable events labelled as slip).
Therefore, when labeling each time steps, we experimentally selected bounds of 3 mm/s and 0.2 rad/s for translational and rotational velocity, respectively, such that almost no slip event would be labelled as `stable.'

\begin{table*}[h!]
\centering
\caption{Distribution of training data by slip type. The cylindrical surface can be oriented in two ways: along the $x$-axis and along the $y$-axis of the fingertip sensor. Marginalized data distributions are listed on the right and bottom of the table.}
\vspace{-1mm}
\label{table:data_dist}
\begin{tabular}{l l *{5}c  }
	\toprule
    \textbf{Slip Type} &  \textbf{Max. Speed} & \multirow{2}{5em}{\centering\textbf{Planar}\\{}} & \textbf{Spherical} & \multirow{2}{6.5em}{\centering \textbf{Cylindrical} ($y$-axis aligned)} & \multirow{2}{6.5em}{\centering \textbf{Cylindrical} ($x$-axis aligned)} & \\
    & & & & & & \\
	\midrule
    \multirow{3}{7em}{\textbf{Translation} (Primary Axes)} &  5 cm/s &      5.3\% & 3.7\% & 3.5\% & 3.6\% & \textbf{16.1\%} \\
    &  7.5 cm/s & 4.5\% & 4.7\% & 3.7\% & 3.7\%  & \textbf{16.7\%} \\
    &  10 cm/s & 4.8\% & 4.7\% & 3.3\% & 3.2\%  & \textbf{15.9\%} \\
    \midrule
    \multirow{3}{7em}{\textbf{Translation} (Oblique Axes)} &  5 cm/s &    4.9\% & 3.3\% & 3.3\% & 3.3\%  &    \textbf{14.8\%}\\
    &  7.5 cm/s & 3.5\% & 4.2\% & 3.5\% & 3.4\%  & \textbf{14.5\%} \\
    &  10 cm/s & 3.9\% & 4.2\% & 3.0\% & 3.1\%  & \textbf{14.3\%} \\
	\midrule
    \textbf{Rotation} &  1 rad/s & 3.8\% & 1.2\% & 1.5\% & 1.2\% & \textbf{7.8\%} \\
    \midrule
    & & \textbf{30.7\%} & \textbf{26.1\%} & \textbf{21.7\%}& \textbf{21.5\%} &  \\
    \bottomrule
\end{tabular}
\vspace{-5mm}
\end{table*}

Our network is trained with tactile measurements collected over surfaces made of smooth, rigid plastic; we focus our study on the following three factors that influence tactile feedback: surface curvature, slip speed, and slip direction. Throughout data acquisition, we maintained a uniform normal contact force.
We chose to collect data for surfaces with three different types of curvature (spherical, cylindrical, and planar), shown in Fig.\ \ref{fig:slip_surfaces}.
The robot moved the sensor at three speeds (0.05 m/s, 0.075 m/s, and 0.1 m/s) in eight different directions to measure translational slip, and rotated the sensor clockwise and counterclockwise at 1 rad/s to measure rotational slip (see Fig.\ \ref{fig:TakkTile_sensor}). This range of motion covers a significant portion of practical slip scenarios.
The maneuvers includes frequent pauses in motion to incorporate static data; our dataset has 143,584 data points belonging to the static class and 122,918 data points belonging to the slip class.
The data in the slip class are evenly distributed across slip speed, slip direction, and surface curvature, as indicated in Table \ref{table:data_dist}.

%%%%%%%%%%%%%%%%%%%%%%%%%%%%%%%%%%%%%%%%%%%%%%%%%%%%%%%%%%%%%%%%%%%%%%%%%%%%%%%%
%% Learning
%%%%%%%%%%%%%%%%%%%%%%%%%%%%%%%%%%%%%%%%%%%%%%%%%%%%%%%%%%%%%%%%%%%%%%%%%%%%%%%%
\section{Learning to Detect Slip}
\label{method:learning}

Prior research has successfully employed deep learning models to extract features from tactile sensor data for slip detection \cite{Zapata-Impata2019,Meier2016,2020_Begalinova_Self-Supervised}.
A very wide range of possible network architectures exist, including TCNs and various recurrent architectures such as LSTMs and ConvLSTMs \cite{Zapata-Impata2019}.
Given indications of higher performance of TCNs on sequence modeling tasks \cite{Bai2018}, we chose to limit our study here to a TCN-based architecture for slip detection.

\subsection{Network Architecture}

As shown in Fig. \ref{fig:TCN_architecture}, the input to the TCN is a stack of pressure values that constitute the last $T_k$ tactile data samples, where $T_k$ is the temporal window size.
The TCN layers are similar to the ones described in \cite{Bai2018}, except that we use layer normalization \cite{ba2016layer} instead of weight normalization \cite{salimans2016weight} and SELU activations \cite{klambauer2017selfnormalizing} instead of the ReLU activations. These architectural alterations were motivated by our extensive experimental tests, which indicated better performance on the tactile dataset.
Although convolutions are only performed along the time dimension, the fully connected layers of the TCN enable multiple temporal inputs to be integrated spatially for classification.

\subsection{Training Details}

The network is trained using the Adam optimizer \cite{kingma2014adam} with a learning rate of 0.002 and a cross-entropy loss function.
We use layer normalization, 20\% dropout (for each layer including the fully connected layers), and a mini-batch of size 256 for network regularization.
At the beginning of each epoch, we equalize the class distribution of the entire training dataset through random undersampling to prevent classification bias.
The dataset was split into training (80\%), validation (10\%) and test (10\%) subsets while maintaining the same proportion of data points in each category, and the network was trained for 800 epochs. A classification accuracy of more than 91\% was achieved within fewer than 100 epochs.

In order to reduce overfitting, we exploited the axial symmetry of the barometer array (see Fig.\ \ref{fig:TakkTile_sensor}) to augment our data. 
Before each epoch, we randomly applied one of four transformations: the identity transform, an $x$-axis flip, a $y$-axis flip, or a 180\textdegree{ }rotation, to every data point with a probability of 25\% each.
After the transformation was applied, we add a small amount of Gaussian noise to the network inputs (with standard deviation equal to 1\% of the barometer range), to ensure robustness to sensor noise. 

\subsection{Temporal Window Size}

\begin{figure}[b!]
    \centering
    \vspace{-4mm}
    \includegraphics[width=0.85\columnwidth]{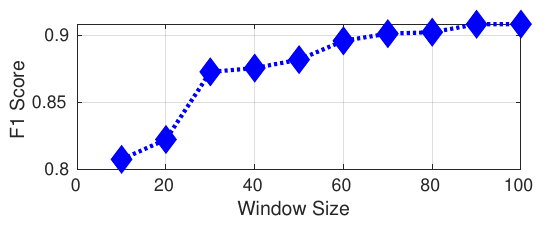}
    \vspace{-2mm}
    \caption{Classification performance (F1-score) versus temporal window size $T_k$ for our proposed TCN architecture.}
    \label{fig:window_size}
\end{figure}

We chose an optimal temporal window size for the input data by training ten different TCNs with window size $T_k$ ranging from 10 to 100 with an increment of 10;
these networks were trained for 400 epochs following the procedure above.
The performance of the networks improved with increasing window size and saturated close to a size of 100 (see Fig. \ref{fig:window_size}).
Inference time is identical for all networks because the temporal convolution operation is parallelized.
A window size of 100 was used in all of the subsequent experiments.
Although the window size is 100, we note that slip can be detected immediately once a new sample arrives.

%%%%%%%%%%%%%%%%%%%%%%%%%%%%%%%%%%%%%%%%%%%%%%%%%%%%%%%%%%%%%%%%%%%%%%%%%%%%%%%%
%% Experiments
%%%%%%%%%%%%%%%%%%%%%%%%%%%%%%%%%%%%%%%%%%%%%%%%%%%%%%%%%%%%%%%%%%%%%%%%%%%%%%
\section{Slip Detection Performance}

We evaluated the performance of our method and compared it with prior work that utilizes contact pressure information to detect slip. Additionally, we characterized the sensitivity of the detector to variations in the factors mentioned in Section \ref{sec:data-collection}.

%%%%%%%%%%%%%%%%%%%%%%%%%%%%%%%%%%%%%%%%%%%%%%%%%%%%%%%%%%%%%%%%%%%%%%%%%
\subsection{Comparison with Frequency-based Methods}
\label{sec: comparison}

We implemented our own versions of the methods in \cite{holweg1996slip} and \cite{Meier2016} and tested them on our dataset.
Holweg et al.\ \cite{holweg1996slip} determine the power spectral density (PSD) of the vibrations induced in a piezoresistive rubber material due to slip; the sum of the high-frequency PSD components is compared against a threshold to infer that slip has occurred.
Meier et al.\ \cite{Meier2016} train a CNN to classify slip events using data from an array of piezoresistive pressure sensors, where the data are first transformed into frequency domain `pressure images'.
We found the optimal PSD threshold by sweeping across various threshold values and evaluating the detector performance on the training dataset.
Table \ref{table:TCN_performance} shows a performance comparison of the above two methods with our TCN-based approach.
While both of the learning approaches demonstrate promising results, slip detection using the PSD threshold exhibits poor performance (as noted in \cite{holweg1996slip} as well).
The TCN outperformed the frequency CNN with an improvement of more than 5\% on each metric  in Table \ref{table:TCN_performance} and predicted the class label with significantly lower variance.
We posit that the difference in performance can be attributed to the fact that Meier et al.\ \cite{Meier2016} enforce a specific transformation of the temporal features in the sensor data, whereas the TCN learns to extract temporal features from the raw sensor outputs.

%%%%%%%%%%%%%%%%%%%%%%%%%%%%%%%%%%%%%%%%%%%%%%%%%%%%%%%%%%%%%%%%%%%%%%%%%%%%%%
\begin{table}[b]
\centering
\vspace{-4mm}
\caption{Performance comparison between TCN and frequency-based methods on our test data. The weighted average of the classes is used to compute each metric.}
\label{table:TCN_performance}
\begin{tabular}{l c c c c }
	\toprule
    \textbf{Method} & \textbf{Accuracy} & \textbf{Precision} & \textbf{Recall} & \textbf{F1-Score} \\ 
    \midrule
    PSD Thresh.\ \cite{holweg1996slip} & 57.4\% & 57.9\% & 57.4\% & 57.5\% \\
    Freq.\ CNN \cite{Meier2016} & 86.0\% & 86.0\% & 86.0\% & 86.0\%\\
    TCN (ours) & \textbf{91.3\%} & \textbf{91.4\%} & \textbf{91.4\%} & \textbf{91.4\%} \\
    \bottomrule
\end{tabular}
\end{table}

%%%%%%%%%%%%%%%%%%%%%%%%%%%%%%%%%%%%%%%%%%%%%%%%%%%%%%%%%%%%%%%%%%%%%%%%%%%%%%
\begin{table*}[t!]
\centering
\caption{TCN performance, measured by F1-score, with variation in slip type, slip speed, slip direction, and surface curvature.} 
\vspace{-1.5mm}
\label{table:TCN_performance_sensitivity}
\begin{tabular}{l l *{4}c  c }
\toprule
     &  \textbf{Max. Speed} & \multirow{2}{5em}{\centering\textbf{Planar}\\{}} & \textbf{Spherical} & \multirow{2}{7.5em}{\centering \textbf{Cylindrical} ($y$-Axis Aligned)} & \multirow{2}{7.5em}{\centering\textbf{Cylindrical} ($x$-Axis Aligned)} &  \multirow{2}{7.5em}{\centering \textbf{Average}\\ (Over Curvatures)} \\
    & & & & & & \\
    \midrule
    \multirow{3}{7em}{\textbf{Translation} (Primary Axes)} &  5 cm/s & 92.5\% & 88.8\% & 78.7\% & 94.0\% & \textbf{89.7\%} \\
    &  7.5 cm/s & 94.9\% & 88.0\% & 90.0\% & 90.3\% & \textbf{91.6\%} \\
    &  10 cm/s & 95.7\% & 84.5\% & 87.8\% & 92.1\% & \textbf{93.1\%} \\
    \midrule
    \multirow{3}{7em}{\textbf{Translation} (Oblique Axes)} &  5 cm/s & 94.0\% & 88.9\% & 87.1\% & 93.2\% & \textbf{91.5\%}\\
    &  7.5 cm/s & 96.6\% & 89.3\% & 93.7\% & 92.2\% & \textbf{93.1\%} \\
    &  10 cm/s & 96.4\% & 88.6\% & 93.5\% & 93.4\% & \textbf{93.8\%} \\
    \midrule
    \textbf{Rotation} &  1 rad/s & 84.6\% & 74.5\% & 84.0\% & 86.1\% & \textbf{81.5\%} \\
    \midrule
    \textbf{Average} (Over Motions) &  & \textbf{94.0\%} & \textbf{88.9\%} & \textbf{88.0\%}& \textbf{91.9\%} & \\
    \bottomrule
\end{tabular}
\vspace{-4mm}
\end{table*}

%%%%%%%%%%%%%%%%%%%%%%%%%%%%%%%%%%%%%%%%%%%%%%%%%%%%%%%%%%%%%%%%%%%%%%%%%%%%%%
\subsection{Detection Sensitivity \& Latency}
\label{sec:performance:sensitivity}

We determined the performance of our detector for possible combinations of slip type, slip speed, slip direction, and surface curvature in order to characterize the sensitivity of the detector.
The classification performance is summarized in Table \ref{table:TCN_performance_sensitivity}, using the F1-score as a metric.

It is clear from the table that the performance of the detector is sensitive to material curvature, slip speed, and slip direction.
Performance appears to be correlated with the number of barometers strongly stimulated by the surface at a given time, which explains the significantly better results for the planar surfaces than for the spherical or $y$-axis-aligned cylindrical surfaces.
The TakkTile fingertip has two rows of MEMS barometers in the $y$ direction as opposed to three rows in the $x$ direction, and this may also explain the improvement when the longitudinal cylinder axis is aligned with the $x$-axis of the fingertip.
We noted a general trend that performance improved as the slip speed increased (before levelling off or decreasing slightly) for almost all curvatures and slip directions. This may be because a higher speed induce larger surface deformations and leads to more prominent temporal features within the one-second input time window.
Performance is also affected by slip direction: slip along the oblique axes of the sensor is better detected than along the primary axes, although the difference is not significant for most slip speeds and curvatures.
The detector produces good results ($>$80\% F1-score) for rotational data as well, with the exception of spherical surfaces.

Once slip occurs, the time taken for the event to be detected is referred to as the \emph{detection latency}.
Keeping this latency low is essential but it ultimately depends on many system-level parameters.
Running the detector in real-time on pre-recorded test data, we measured an average network inference time of 21 ms and an average latency of 134 ms, which is higher than the $\sim$75 ms latency of the human response to slip events \cite{johansson1984roles}.
We found that, on average, 11.3 additional sensor readings were required for slip to be detected following inception, which accounts for 113 ms of the overall latency.
For our prototype, no direct efforts were made at the system level to reduce this latency.
Latency could be greatly reduced with higher-frequency signal processing hardware---the MEMS barometers can operate at a sampling rate of up to 400 kHz, while the TakkTile sensor board is limited to providing data at 100 Hz only.

\begin{figure}[b]
    \centering
    \vspace{-4mm}
    \includegraphics[trim=0 20 0 80, clip,width=0.725\columnwidth]{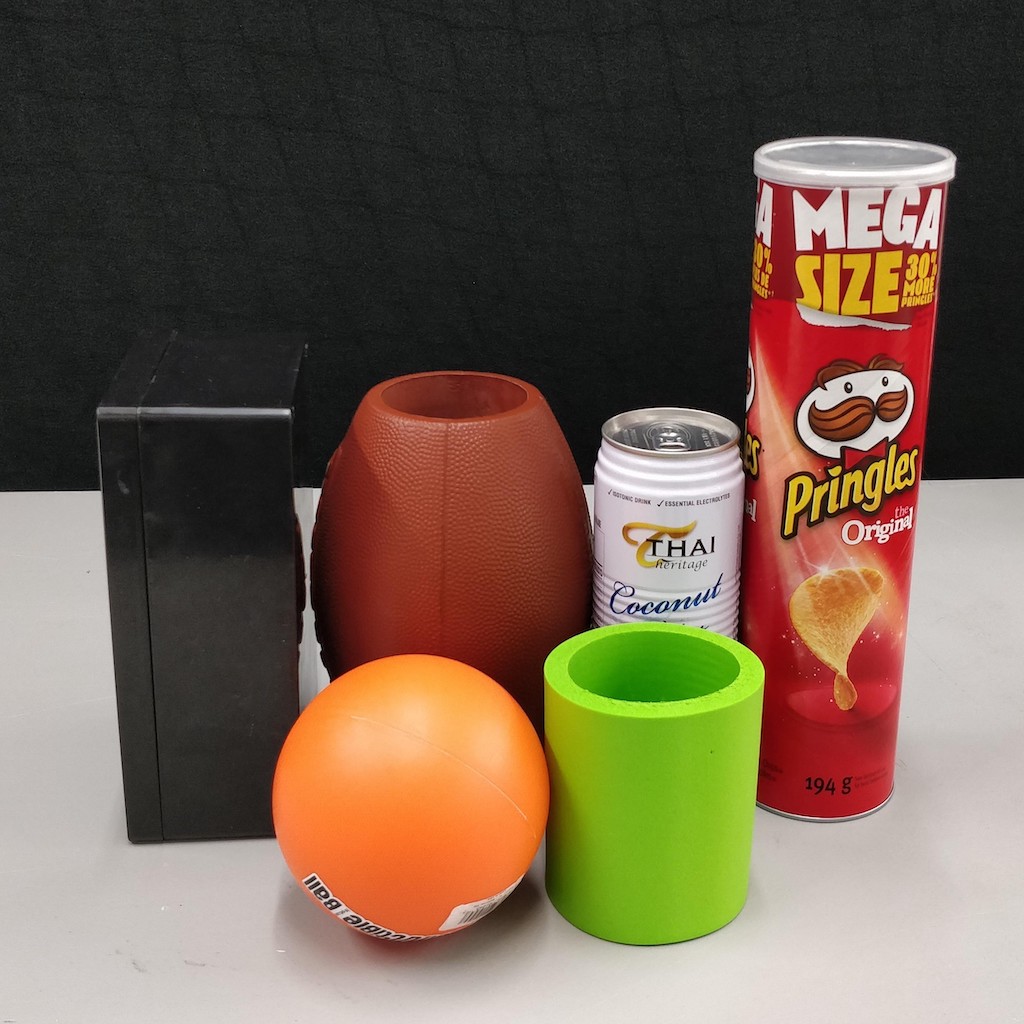}
    \caption{Objects used for our robot experiments. From left to right, the objects and  surface properties are:  \textit{plastic box} (planar, smooth, rigid), \textit{plastic ball} (spherical, smooth, rigid), \textit{football sleeve} (spherical, rough, deformable), \textit{foam sleeve} (cylindrical, smooth, deformable), \textit{metal can} (cylindrical, rough, rigid), and \textit{cardboard can} (cylindrical, smooth, rigid).}
    \label{fig:test_objects}
    \vspace{-1mm}
\end{figure}

\section{Robot Experiments}
\label{sec: robot_experiments}

One of our objectives was to determine how well our detector generalizes to common real-world slip-detection tasks. 
We tested the generalization capability of our network model on the following two manipulation tasks: slip detection for an in-hand object under an externally-applied impulsive force, and slip detection while lifting an object with insufficient grasping force.
In our experiments, if slip was detected for two consecutive time steps, a slip event was registered and a response was executed, if needed.
The test objects for the experiments, shown in Fig.\ \ref{fig:test_objects}, were purchased off-the-shelf and were selected with the intention of varying the properties of the contact surfaces, such as curvature, roughness and deformability.
While our network was trained solely on 3D-printed rigid ABS plastic, we intentionally tested on objects with dissimilar surface properties to gauge the limits of our detector.

\subsection{Mallet Tap Test}

\begin{figure}[t]
\centering
\includegraphics[width=0.8\columnwidth]{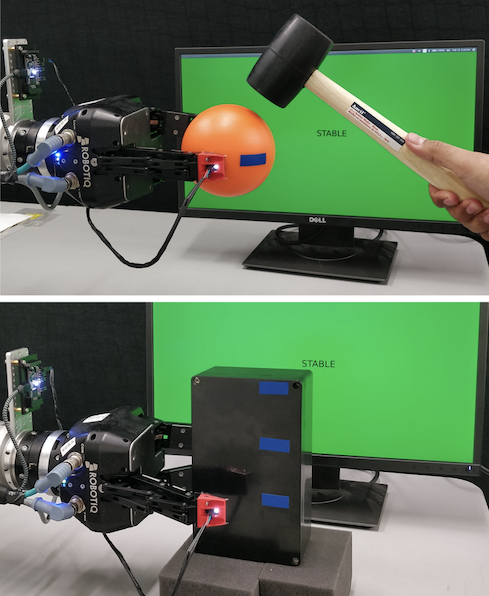}
\caption{Mallet test for one trial with the plastic ball (top) and lift test for one trial with the plastic box (bottom). The blue tape on the objects marks the initial grasp location. The box has additional blue markings 6 and 12 cm above the initial grasp location for reference.}
\label{fig: object_lift_test}
\vspace{-5mm}
\end{figure}

The goal of this experiment was to evaluate the performance of the slip detector for in-hand objects grasped with a constant force.
As the gripper held an object above the test table, the object was manually tapped using a 16 oz rubber mallet with enough force to induce slip, but not to cause complete grasp failure.
The force applied by the gripper to each object was 10 N, which is the minimum force that the device can be programmed to exert.
The setup for this test is shown in Fig.\ \ref{fig: object_lift_test}; the LCD monitor in the background displayed the slip registration status.

Each trial involved tapping the object and determining (according to the network output) whether slip had been detected. We then confirmed that displacement of the object had occurred by measuring the position of the blue tape (shown in Fig.\ \ref{fig: object_lift_test}) relative to its initial position at the start of the trial (using captured test images).
A trial was deemed successful if slip was registered after the object moved upon being tapped and the status returned to the nominal value after the object stopped moving.

We conducted 20 trials per object, with an equal number of taps on the top and on the side of the object.
The slip detector achieved an average accuracy of 87.5\%. 
We noted that detection performance was best for smooth objects, but the results did not display a consistent trend in terms of curvature.
The worst performance (80\%) was obtained with the foam sleeve, likely due to its deformable surface.
However, as shown in Table \ref{table:experiment_results}, the slip detector performed well overall under constant load conditions despite variations in material, curvature, deformability, and smoothness of the object surfaces.
Since contact was maintained at all times, this test verified that true object slip was being detected, rather than simply the detection of a uniform reduction in force across the tactile sensor array (i.e., a sudden `air gap' upon reduction in the grasping force). 

%%%%%%%%%%%%%%%%%%%%%%%%%%%%%%%%%%%%%%%%%%%%%%%%%%%%%%%%%%%%%%%%%%%%%%%%%%%%%%%%%

\subsection{Object Lift Test}

When lifting an object, humans intuitively change their grasping force based on the weight of the object.
If a mismatch exists between the perceived and the actual weight, slip may occur.
We replicated this scenario with our gripper and arm using the same set of objects shown in Fig.\ \ref{fig:test_objects}.
Since the Robotiq gripper is not built for fine finger control, it was challenging to find gripper configurations that maintained contact with the object while generating insufficient grasping force.
To solve this problem, we added weights to the lighter objects to increase the likelihood of slip.
Fig.\ \ref{fig: object_lift_test} shows the setup used for this experiment. 
To ensure repeatability, we used a predetermined gripper configuration for each object and kept the initial grasping position consistent throughout the experiments. 
Once the gripper fingers reached their predetermined positions, the slip detector was initialized---this was followed by a lifting motion.

An experimental trial was considered successful if slip was correctly detected and compensated for, that is, if the object was stably grasped and lifted.
To compensate for slip, we commanded the gripper to close as quickly as possible while not exceeding an applied force of 10 N.
For the longer objects (i.e., the cardboard can, the plastic box, and the metal can), successful trials were split on the basis of the distance between the initial and the final grasp position---distance thresholds of 6 cm and 12 cm were used to define these two categories.
We conducted 20 trials for each object; the experimental results are shown in Table \ref{table:experiment_results}. 
As a general trend, the success rate is greatest for longer objects, which is due to the reaction time: there is more time to `catch' longer objects.
Similarly, a high success rate was achieved with the football sleeve, which we attribute to its longer body and rough surface texture.

\begin{table}[b]
\vspace{-4mm}
\centering
\caption{Slip detection results for two real-world experiments.} \label{table:experiment_results}
\vspace{-1mm}
\begin{minipage}{\columnwidth}
\renewcommand{\thempfootnote}{\arabic{mpfootnote}}
\begin{center}
	\begin{tabular}{ l  l  l}
	\toprule
    \textbf{Object} & \textbf{Mallet Tap} & \textbf{Object Lift} \\
	\midrule
    Plastic Ball & 90\% & 35\% \\
    Plastic Box & 85\%  & 100\%\footnotemark[3] \\
    Cardboard Can & 90\%  &  85\% (45\%\footnotemark[3] + 40\%\footnotemark[4]) \\ 
    Football Sleeve & 90\%  & 85\% \\
    Foam Sleeve & 80\%  & 35\% \\
    Metal Can & 90\%  &  95\% (75\%\footnotemark[3] + 20\%\footnotemark[4]) \\
    \midrule
    \textbf{Avg. Success} & \textbf{87.5\%} & \textbf{72.5\%} \\
    \bottomrule
\end{tabular}
\end{center}
\vspace{-4mm}
\hfill
\begin{minipage}{0.925\columnwidth}
\footnotetext[3]{Final grasp within 6 cm of the initial grasp position.}
\footnotetext[4]{Final grasp between 6-12 cm of the initial grasp position.} 
\end{minipage}
\end{minipage}
\end{table}

%%%%%%%%%%%%%%%%%%%%%%%%%%%%%%%%%%%%%%%%%%%%%%%%%%%%%%%%%%%%%%%%%%%%%%%%%%%%%%%%
%% CONCLUSION
%%%%%%%%%%%%%%%%%%%%%%%%%%%%%%%%%%%%%%%%%%%%%%%%%%%%%%%%%%%%%%%%%%%%%%%%%%%%%%%%
\section{Conclusions and Future Work}
\label{sec:conclusion}

We have presented a learning-based method for slip detection using inexpensive barometric tactile sensors.
We collected training data with variations in surface curvature, slip speed, slip direction, and slip type, and assessed the sensitivity and robustness of our method to these factors.
To demonstrate the accuracy and performance of our slip detection algorithm, we compared its performance with an existing classical method and a learning-based method.
Our TCN outperformed these methods with an accuracy greater than 91\% on our dataset.
Finally, we evaluated the performance of the proposed slip detector on two real-world robotic tasks, using objects with different surface properties such as curvature, roughness, and deformability.
The detector displayed high sensitivity to slip type and surface curvature, while being relatively robust to slip speed and direction.
We believe that detection performance can be further improved by increasing the spatial density of the MEMS barometers on each fingertip, while the latency could be reduced by sampling the sensors at a higher rate.

As future work, we would like to investigate the benefits of employing multiple TakkTile fingertips and a palm sensor for slip detection during manipulation.
Similarly, an ability to estimate the type and direction of slip would be valuable.
Finally, we would like to examine the use of different barometric sensor configurations for tactile sensing, including flexible `skin,' similar to \cite{koiva2020barometer}.

%%%%%%%%%%%%%%%%%%%%%%%%%%%%%%%%%%%%%%%%%%%%%%%%%%%%%%%%%%%%%%%%%%%%%%%%%%%%%%%%
%% REFERENCES
%%%%%%%%%%%%%%%%%%%%%%%%%%%%%%%%%%%%%%%%%%%%%%%%%%%%%%%%%%%%%%%%%%%%%%%%%%%%%%%%
\bibliographystyle{IEEEcaps}
\bibliography{abbrevs,refs}

\end{document}